\documentclass{article}

\usepackage{PRIMEarxiv}

\usepackage[utf8]{inputenc} % allow utf-8 input
\usepackage[T1]{fontenc}    % use 8-bit T1 fonts
\usepackage{hyperref}       % hyperlinks
\usepackage{url}            % simple URL typesetting
\usepackage{booktabs}   % nice looking tables
\usepackage{siunitx}% professional-quality tables
\usepackage{amsfonts}       % blackboard math symbols
\usepackage{nicefrac}       % compact symbols for 1/2, etc.
\usepackage{microtype}      % microtypography
\usepackage{lipsum}
\usepackage{fancyhdr}       % header
\usepackage{graphicx}       % graphics
\graphicspath{{media/}}     % organize your images and other figures under media/ folder
\usepackage{numprint} %in preamble
\usepackage{array}
\newcolumntype{C}[1]{>{\centering\arraybackslash}m{#1}}
\title{Dynamic Temporal Reconciliation by Reinforcement learning
}

\author{
  Himanshi Charotia \\
  Mastercard, AI Garage \\
  Gurugram\\
   \And
  Abhishek Garg \\
  Mastercard, AI Garage \\
  Gurugram\\
    \And
    Gaurav Dhama \\
  Mastercard, AI Garage \\
  Gurugram\\
    \And
    Naman Maheshwari \\
  Mastercard, AI Garage \\
  Gurugram\\
}

\begin{document}
\maketitle

\begin{abstract}
Planning based on long and short term time series forecasts is a common practice across many industries. In this context, temporal aggregation and reconciliation techniques have been useful in improving forecasts, reducing model uncertainty, and providing a coherent forecast across different time horizons. However, an underlying assumption spanning all these techniques is the complete availability of data across all levels of the temporal hierarchy, while this offers mathematical convenience but most of the time low frequency data is partially completed and it is not available while forecasting. On the other hand, high frequency data can significantly change in a scenario like the COVID pandemic and this change can be used to improve forecasts that will otherwise significantly diverge from long term actuals.

We propose a dynamic reconciliation method whereby we formulate the problem of informing low frequency forecasts based on high frequency actuals as a Markov Decision Process (MDP) allowing for the fact that we do not have complete information about the dynamics of the process. This allows us to have the best long term estimates based on the most recent data available even if the low frequency cycles have only been partially completed. The MDP has been solved using a Time Differenced Reinforcement learning (TDRL) approach with customizable actions and improves the long terms forecasts dramatically as compared to relying solely on historical low frequency data. The performance of the proposed Reinforcement Learning system has been evaluated on stock exchange NIFTY 50 dataset\cite{nifty50}. The result also underscores the fact that while low frequency forecasts can improve the high frequency forecasts as mentioned in the temporal reconciliation literature (based on the assumption that low frequency forecasts have lower noise to signal ratio) the high frequency forecasts can also be used to inform the low frequency forecasts. However, this is not achievable without accepting the fact that the dynamics of the system in practical scenarios can change significantly in a short amount of time and the temporal reconciliation model has to account for these uncertainties. 
\end{abstract}

% keywords can be removed

\section{Introduction}
Many real-life decision problems involve multiple time horizons. For example, in energy planning expected demand has to be forecasted for the upcoming hours, the next day, the next week, and even longer horizons \cite{JEON2019364, Taeib2020}. Another example is inventory management where short term(monthly) forecasts are needed for next month's stocking. Whereas it is often beneficial to take into account long term forecasts for taking decisions related to establishing contracts of purchasing a product \cite{seeger2016_03255088, seeger2017approximate}. These decisions are made one-two year in advance using long term forecasts (1-2 years). This is to make sure that decisions made now have a positive impact on future possibilities.

In this context, temporal aggregation and reconciliation techniques have been found to be useful in improving forecasts among multiple horizons \cite{Amemiya_and_yu, tiao, silverstreni}. Different temporal aggregations can reveal important information about the underlying data-generating process. When temporal aggregation is applied to a time series, it can strengthen or attenuate different features \cite{Januschowski2019}. Temporal hierarchies forecasts for different horizons can be made with generating forecasts independently for time series at each level by different (simple) methods. These forecasts produced by different approaches and based on different information are most likely incoherent. Reconciliation is necessary because optimal decision-making requires coherent forecasts.

Thus, the main focus while forecasting hierarchical time
series is to utilize the information available across all levels
of a given hierarchy for producing coherent forecasts.
Existing work in hierarchical forecasting deploys methodology in which first base forecasts are independently generated for each time series in the hierarchy and then combination/revision step is done
in a post-processing step to ensure coherence. However , all the methods used in the industry or literature rely on the availability of complete data across all levels of the temporal hierarchy. While this offers mathematical convenience it is often not available to the decision maker using the forecasts. Due to the dynamic and stochastic nature of time series, it is very difficult to precisely estimate future changes (significant trend or mean shift) and to adjust different horizon forecasts based on these changes. An example can be a significant trend or mean shift in high frequency patterns at a daily or a weekly level such as those encountered in the COVID pandemic. There are high chances that the long term actuals will significantly diverge from the forecasts. All the existing methods lead to static temporal reconciliation and thus will perform poorly under dynamic scenarios. 

%\subsection{Why RL is best suited}
Reinforcement learning especially deep reinforcement learning (DRL) has been successfully applied in various fields including AlphaGo \cite{silver2017mastering}, ATARI games \cite{DBLP:journals/corr/abs-1207-4708} and robotics \cite{rl_robotics} . Reinforcement learning is a task-independent learning
scheme. It is suitable for problems where there is no
supervised information but only feedbacks from an external
environment. The problem of informing forecasts among hierarchies for reconciliation without prior/complete knowledge of data distribution can be categorized as the same. Furthermore, reinforcement learning is a data-driven approach that is able to capture complex changing dynamics in the data and is well equipped to overcome the deficiencies of the existing methods.

%\subsection{Framework}
We propose a dynamic reconciliation method whereby we formulate the problem of informing low frequency forecasts based on high frequency actuals as a Markov Decision Process (MDP) allowing for the fact that we do not have complete information about the dynamics of the process. This allows us to have the best long term estimates based on the most recent data available even if the low frequency cycles have only been partially completed. The MDP has been solved using a Time Differenced Reinforcement learning (TDRL) approach and improves the long terms forecasts dramatically as compared to relying solely on historical low frequency data. The temporal difference is an agent learning from an environment through episodes with no prior knowledge of the environment. We have designed our own action and reward function. Based on the factors described above, the contributions of this
paper can be summarized as below:
{\begin{itemize}
\item To the best of our knowledge, our work presents the first RL approach for the hierarchical reconciliation problem.
\item Dynamic reconciliation framework which works on partial data also.
\item A tunable action design function that tries to incorporate forecast change with tolerance while reconciling the forecast which ensures that produced reconciled forecast do not deviate from expected values.
\end{itemize}}
The rest of this paper is organized as follows. Section 2 introduces
the reinforcement learning basics. In Section 3, we describe the detailed methodology of the proposed TDRL approach towards the hierarchical time series reconciliation and give a detailed analysis on how to design actions and \(\epsilon\)-greedy policy. We report and discuss experimental results in Section 4. We talk about future directions and conclude in Section 5.
% add more details about proposed model's working 

\section{Background}
\label{sec:background}

%\lipsum[4] See Section \ref{sec:headings}.

\subsection{Hierarchical Reconciliation}

Temporal hierarchies for forecasting  are constructed for any time series by means of non-overlapping temporal aggregation. Such aggregation typically leads to tree structure but need not necessarily. For example, grouped \cite{HYNDMAN201616}, temporal \cite{ATHANASOPOULOS201760}, and cross-temporal aggregations \cite{SPILIOTIS2020114339} can be alternative aggregation paths.

Consider a multi-level hierarchy \(\mathbf{Y}_t \in \mathbb{R}^{n} \) at time \textit{t}  having \(t = 1, . . . , T \). Here \(y_{t,i} \in \mathbb{R} \) is the value of the \textit{i}-th (out of n) univariate time series. The index \textit{i} denotes level of hierarchy. Level 0 (\textit{i}=0) denotes the completely aggregated series, level 1 the first level of disaggregation, down to level K containing the most disaggregated time series.

We refer to the time series at the leaf nodes of the hierarchy as bottom-level series and rest of the series can be termed as aggregated series. We also call the forecasts for all time series in the hierarchy generated without any reconciliation approach as base forecasts denoted by \(\mathbf{\hat{Y}}_t\)(not to be confused with bottom-level). 
We can split the vector of all series \(\mathbf{Y}_t\) into \textit{m} bottom entries and \textit{r} aggregated entries where  n = r + m. 
And \(\mathbf{Y}_t = [u_\textit{t} \hspace{0.1cm} v_\textit{t}]^\top \) with \(u_\textit{t} \in \mathbb{R}^{r}\) and \(v_\textit{t} \in \mathbb{R}^{m}\). 
Now, the aggregation matrix \(S \in \{0,1\}^{n\times m} \) is defined and the \(\mathbf{Y}_t, v_\textit{t}\) and \(S\) satisfy
\begin{equation}
    \mathbf{Y}_t = Sv_\textit{t}
\end{equation}
%\mathbf{Y}_t = Sv_\textit{t}

Existing approaches for generating a coherent forecasts for a hierarchical time series follow a two-step procedure: (i)generate h-step-ahead forecast for each time series independently to obtain base forecasts \(\hat{Y}_{T+h} \) and
(ii) produce revised h-step-ahead forecasts \(\tilde{Y}_{T+h} \) through reconciliation given by equation below:
\begin{equation}
    \tilde{Y}_{T+h} =SP \hat{Y}_{T+h}
\end{equation}

for some appropriately chosen matrix P of order \(m \times n\). P matrix is used to extract and combine the relevant elements of the base forecasts \(\hat{Y}_{T+h} \). 

\subsection{Temporal difference Reinforcement learning}
Reinforcement learning (RL) is a machine learning approach inspired
by behaviorist psychology. In RL, an agent interacts with
its environment by sequentially taking actions, observing consequences,
and altering its behaviors to maximize a cumulative
reward. RL is usually modeled as a MDP which consists of a state space \( \mathcal{S} = \{s\} \), an action space
\( \mathcal{A} =  \{a\} \), state transition dynamics \(  \mathcal{T: } { \mathcal{S} \times \mathcal{A}}\rightarrow \mathcal{P (\ S )\ }  \)
where \(\mathcal{P (\ S )\ }  \) is the set of probability measures on S, an immediate reward
function \( r { : \mathcal{S} \times \mathcal{A}}\rightarrow \mathbb{ R } \), and a discount factor \( \gamma \in [\ 0, 1 ]\ \) . A policy, denoted by \(  \mathcal{\pi : }  \mathcal{S} \rightarrow \mathcal{P (\ A )\ }  \) where \(\mathcal{P (\ A )\ }  \) is the set of probability measures on \(\mathcal{A }  \), fully defines the behavior of an agent. The agent
uses its policy to interact with the environment and gives a trajectory
of states, actions, and rewards \( {s_1}, a_1, r_1, ..., s_T , a_T , r_T \)  (T = \(\infty \)
indicates a infinite horizon MDP and otherwise an episodic one)
over \( { : \mathcal{S} \times \mathcal{A}\times \mathbb{R}} \). The cumulative discounted reward constitutes
the return \( \mathcal{R} =\sum_{t=1}^{T} {\gamma^ {t-1} }r_t\) . The agent’s goal is to learn an optimal
policy \(\pi^{\ast } \)  that maximizes the expected return from the start state.
\(\pi^{\ast} = arg {max_{\pi}} \mathbb{E}[\ {R |\pi} ]\  \). A common method for learning optimal policy or optimal state-value function is temporal difference (TD) learning, which estimates the value of a state by bootstrapping from the value estimates of successor states using Bellman-style equations. TD methods work by updating the state-value estimates to reduce the TD error, which describes the difference between the current estimate of the state-value and a new sample obtained from interacting with the environment.

\section{Related Work}
Top-down and bottom-up approaches have traditionally been used to produce coherent forecasts for a hierarchy. In Top-down, forecasts are generated at the top level for the time series and then dis-aggregated down all the way to the bottom level. Whereas for Bottom-up, forecasts are generated at the most granular level and then aggregated up \cite{ATHANASOPOULOS2009146, Gross}. In both of these methods, the generation of forecasts for the entire hierarchy is dominated by a single level of aggregation where the forecasts are produced, ignoring information at all other levels.
%Ensembling of these approaches are presented in (Hollyman et al., 2021).
To optimally combine forecasts from all the series of the hierarchy, \cite{hyndman_shang2011.03.006} proposed the use of ordinary least-squares (OLS) estimator after formulating the forecast reconciliation problem for a structural hierarchy as a linear regression model. In the regression, the independent base forecasts are modeled as the sum of the expected values of the future series and coherency error. \cite{HYNDMAN201616} suggested using weighted least squares (WLS), taking account of the variances on the diagonal of the covariance matrix of aggregation error but ignoring the off diagonal covariances. Later, \cite{hyndman_wickra2018} considered the generalized least-squares (GLS) estimator using \( P = {(S^\top {W_h}^{-1} S)}^{-1} {(S^\top {W_h}^{-1})}\), where \( W_h\) is the covariance matrix of the \textit{h}-period-ahead forecast errors \( \hat{\varepsilon}_{T+h}= {Y}_{T+h}-\hat{Y}_{T+h} \).
MinT found that the incorporation of correlation information into the reconciliation procedure to be beneficial for forecast accuracy, when combined with a simple shrinkage estimator. The advantages of the MinT approach are that its revised forecasts are coherent by construction and it uses information from all levels of hierarchy simultaneously. Disadvantages are the strong assumption of base forecasts to be unbiased.
\cite{ATHANASOPOULOS201760} showed that it is possible to use the reconciliation framework proposed by \cite{hyndman_shang2011.03.006} to produce coherent forecasts by representing temporally aggregated series as hierarchical time series. The issue of getting coherent forecasts along both cross-sectional and temporal dimensions (i.e., cross-temporal coherency) has been dealt with by \cite{YAGLI2019391, SPILIOTIS2020114339, KOURENTZES2019393} and \cite{difonzo2020crosstemporal}.

\section{Method}
\subsection{Problem Definition}

For this paper, we will use only 2-level of hierarchy. Level-0 denotes monthly aggregated data and Level-1 denotes its disaggregation into daily data.  
The framework will take as input aggregated (monthly) forecasts, bottom level (daily) forecasts \( \hat{Y}\) which are provided to any reconciliation approach. In addition to this, the framework will have access to ongoing high frequency actual. High frequency actual can help the agent to learn the daily shares and will help the agent to model ongoing dynamics in the data. 

We consider an episodic MDP with a discount factor \( \gamma =1 \) where an episode(typically one month) starts with a low frequency forecast. From the external model, the agent also gets a share of each day in a month. The agent is given high frequency forecast \( \hat{y}_t \) at timestamp t, the agent takes an action of increasing or decreasing forecast per day to capture the daily variation/fluctuation. The main goal of the agent is to update monthly forecasts based on lower level forecasts and actual. This is being done using reward function R. 
%This seems like a trivial problem to solve when lower level data is true representative of higher level but several cases may arise such as mean or trend shift and in those cases the long term actual will not align with the forecasts. %
The goal of the agent is to generate reconciled low frequency forecasts based on the high frequency actual share seen per day. More specifically, the core elements of the MDP are further explained as follows:
{\begin{itemize}
\item \(\mathcal{S}\): States in the environment. The state here represents the number of days present in a month. This will help the agent to keep track of how many days have passed in a month and the corresponding daily share. State \(s_t\) is represented as : 1) \textit{t} : the current day of the month, 2) State value will be remaining monthly total to be adjusted.

\item \(\mathcal{A}\): It constitutes a set of actions present to the agent in a state S. We have designed our own set of actions which will be discussed in the later section.
\item \(\mathcal{T}\) : Transition function of MDP. We have used \( \epsilon -\) greedy policy which mixes two behavior of switching between random and Q policy using the probability hyperparameter  \( \epsilon \). 
\item \(\mathcal{R} \): A reward \(r_t\) is a scalar which measures the goodness of the action \(a_t\)  taken by the agent in the state \(s_t\). Reward \(r_t\) at each time step \textit{t} will be high frequency actual encountered for that day. 
\item \( \gamma \)  : We set reward discount factor \( \gamma \) = 1 
\end{itemize}}

\subsection{Action Design}
Actions help the agent to interact with the environment and get feedback.  
As the main goal is the adjustment of the forecast, we have defined actions accordingly. The Agent has to perform an action of increase or decrease in forecast per day to capture the daily variation/fluctuation. But how much to increase or decrease from daily level forecast can have infinite values. To make this action space discrete, we have introduced the tolerance level \( \varepsilon \) parameter which helps in bucketing the action space.
In any state \(s_t\), agent can increase or decrease the high frequency forecast based on tolerance level \( \varepsilon \). Lets take an example of 3 actions, then for any time step \textit{t} the Agent can make updated daily level forecast to be
{\begin{enumerate}
\item  \(  \geq \hat{y_t} + \varepsilon \) 
\item \( = \hat{y_t} \pm \varepsilon \)
\item \(  \leq \hat{y_t} - \varepsilon \)
\end{enumerate}}
The tolerance level \( \varepsilon \) parameter can be tuned according to the problem at hand and helps in deciding how much fluctuation from the daily level forecast is permitted to the agent. To illustrate the concept, lets consider \( \varepsilon\) =5\$ and daily level forecast value at time \textit{t} to be 30\$. Then the agent can take one of the 3 actions of increasing/decreasing forecasts defined as:
{\begin{enumerate}
\item  \(  \geq 30 + 5 \) 
\item \( = 30 \pm 5 \)
\item \(  \leq 30 - 5 \)
\end{enumerate}}

\subsection{Policy}
 Epsilon-Greedy is a method for selecting actions that balance both exploration and exploitation by choosing between exploration and exploitation randomly. The epsilon-greedy, where  \( \epsilon \) refers to the probability of choosing to explore. A policy is \( \epsilon \) -greedy with respect to an action-value function estimate Q if for every state,
{\begin{itemize}
\item with probability \(1-\epsilon \) , the Agent selects the greedy action, and
\item with probability \( \epsilon \) , the Agent selects an action uniformly at random from the set of available (non-greedy and greedy) actions.

\end{itemize}}
The probability of selecting non-greedy actions increases with larger value of \( \epsilon \). 
To construct a fixed policy \( \pi \)  that is \( \epsilon \) -greedy with respect to the current action-value function estimate Q , we have set the policy as
\begin{equation}
  1-\epsilon + \frac{\epsilon}{|A(S)|} 
\end{equation}
if action \textit{a} maximizes Q ( s , a ). Else
\begin{equation}
  \frac{\epsilon}{|A(S)|} 
\end{equation}
for each \(\textit{s} \in S \) and \(\textit{a} \in A( s ) \).

\begin{figure}[h]
\centering
\includegraphics[width=15cm]{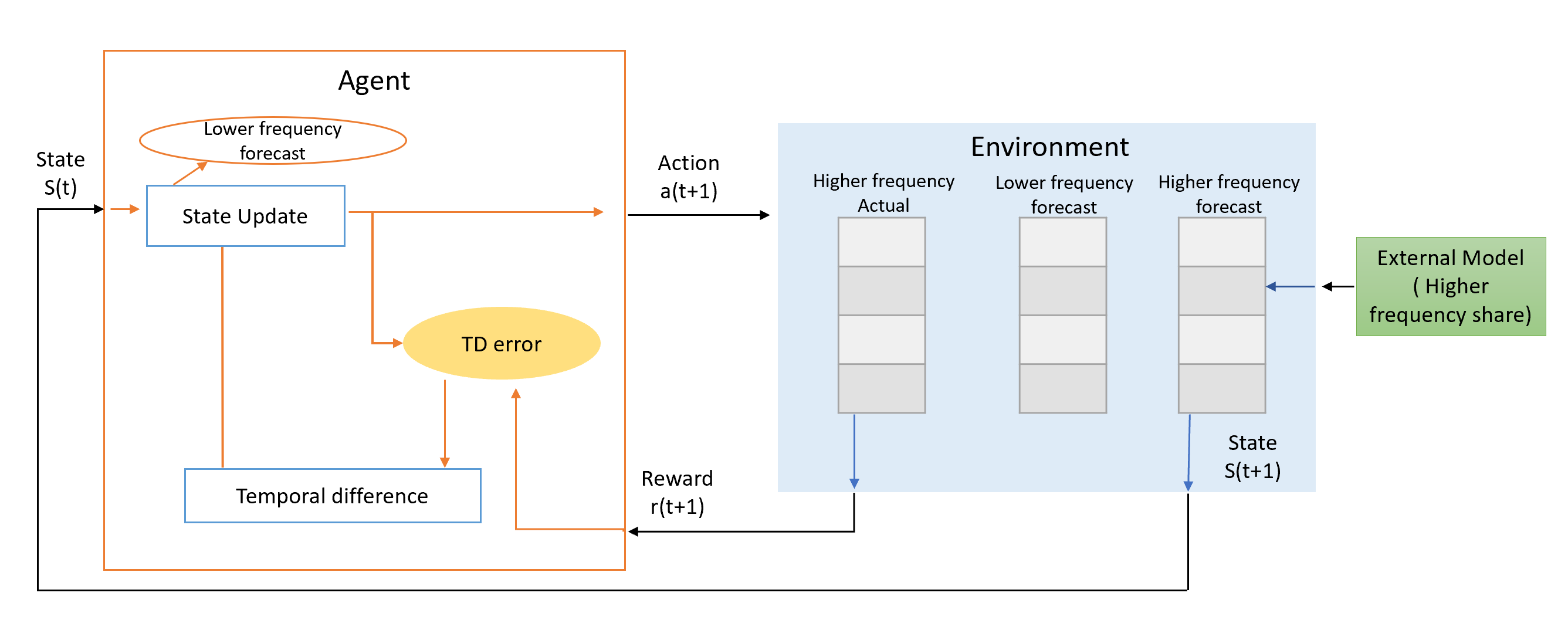}
\caption{ Environment design for agent training}
\label{fig:rl_env}
\end{figure}

\subsection{State-Value Update}
For finding the optimal state-value function, we have used SARSA variation of Temporal Difference-(0) Algorithm. TD(0) improves upon the drawbacks of the Monte Carlo (MC) method in which the agent has to wait until the end of an episode to obtain the actual return (experienced return) before it can update and make any improvements to the value function estimate. TD(0) updates the state value function at every time step \textit{t} without waiting for the end of the episode. For finite-state MDP under a policy \(\pi\), the update rule is given by: 
\begin{equation}
    V(S_t) \leftarrow V(S_t)+ \alpha(R_{t+1}+\gamma V(S_{t+1}-V(S_t))
\end{equation}
where \( V(S_t) \) denotes is state value function for state \( S_t \), \( \alpha\) is step-size parameter, \( \gamma\) is discount factor and \( R_{t+1}\) is immediate reward recieved at time step \textit{t+1}.  
\begin{figure*}
\centering
\includegraphics[width=16cm,height=4cm]{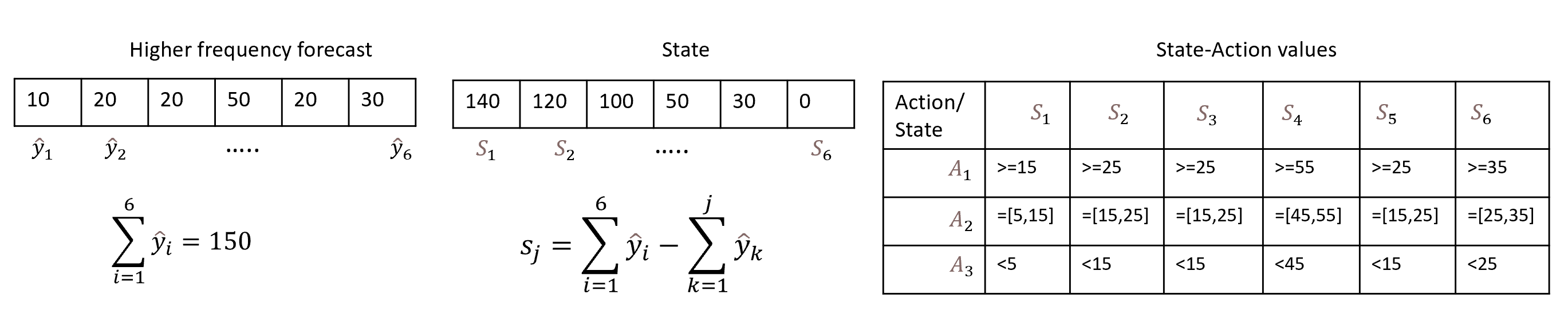}
\caption{ An example of state-action initialization for a time series. \( \hat{y_1}\) to \( \hat{y_6}\) are forecasted values of a time series observed at \textit{t} \( \in \{{1,....,6}\}\) respectively. Considering time step \textit{t}=2, \(S_2\) will be initialized as 150-(10+20)=120. Three actions for \(S_2\) are defined based on \( \epsilon\)=5 and \(\hat{y_2}\)=20.    }
\label{fig:rl_example}
\end{figure*}
\begin{figure*}
\centering
\includegraphics[width=16cm,height=6cm]{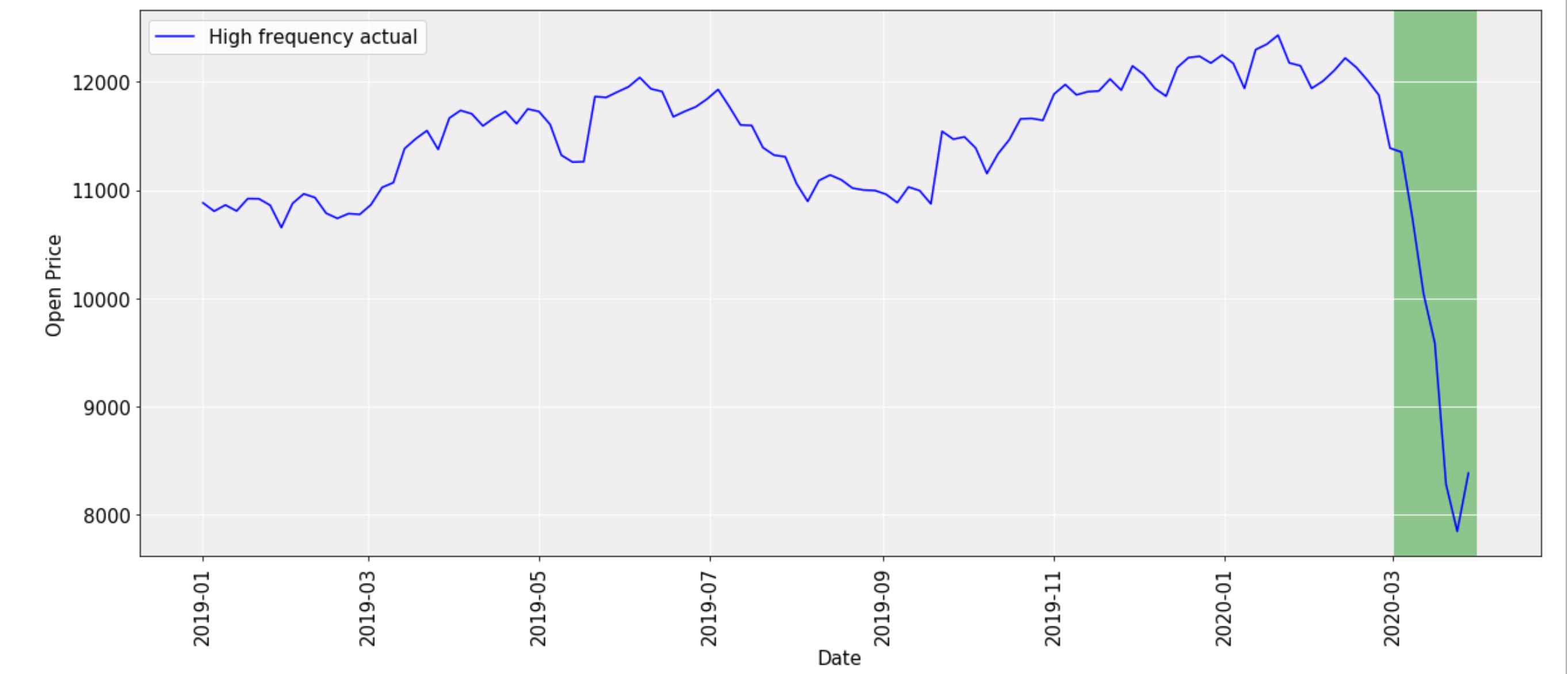}
\caption{High frequency dataset starting from January 2019 to March 2020. X-axis denotes dates and Y-axis denotes open price value of Nifty 50 index. Test period is highlighted in green.}
\label{fig:rl_actual}
\end{figure*}   

\subsection{Reinforcement learning to Reconciliation}

Putting them together, we present our Dynamic Temporal Reconciliation (DTR) framework as given in Fig \ref{fig:rl_env}. The framework is built using TD(0) which takes as input high frequency forecasts, where the state values are first initialised as difference between forecasted low frequency total and cumulative high frequency forecasts seen till time step \textit{t} for every \textit{t} \( \in \{{1,....,n}\}\). Action value function are initialised using the three actions defined in Section 3.2  based on forecast for every \textit{t} \( \in \{{1,....,n}\}\) for the construction of corresponding \(\epsilon\)-Greedy policy. See Fig. \ref{fig:rl_example} for an illustration. 
Based on the \(\epsilon\)-greedy policy, the agent takes an action \(a_t \in A\)  under state \(s_t \in S\). Here our reward function \( r_t \) is not directly beneficial for the agent at each step but combined with the value update function, it helps in achieving the overall goal of the agent.
\begin{table*}
\npdecimalsign{.}
\nprounddigits{2}

\caption{Performance comparison for reconciled task for different settings for tolerance level \(\varepsilon\) with fixed \(\epsilon\)=0.05 (RMF: Reconciled Monthly Forecast) \\
Note : the columns represent three values ( RMF / \textbf{\( MAPE\)} / \textbf{\( \%_f\)} )}
\label{table : results }
\begin{center}
\setlength\extrarowheight{0.8pt}
\begin{tabular}{|c|c|c|c|c|c|}
\hline
\textbf{Date}  & \textbf{Actual} & \textbf{Forecasted} & \begin{tabular}[c]{@{}c@{}}\textbf{RMF} \\ \(\varepsilon\) = \textbf{0.1} \end{tabular} & \begin{tabular}[c]{@{}c@{}}\textbf{RMF} \\ \(\varepsilon\) = \textbf{0.2}\end{tabular} & \begin{tabular}[c]{@{}c@{}}\textbf{RMF} \\ \(\varepsilon\) = \textbf{0.3}\end{tabular} \\ \hline
01/03/20      & 11386           & 11354               & 367706 / 25\% / 0\%                                                   & 367706 / 25\% / 0\%                                                   & 364148 / 24\% / 1\%                                                   \\
02/03/20      & 11387           & 11472               & 368892 / 25\% / 0\%                                                   & 367706 / 25\% / 0\%                                                   & 367706 / 25\% / 0\%                                                   \\
03/03/20      & 11218           & 11513               & 367706 / 25\% / 0\%                                                   & 367706 / 25\% / 0\%                                                   & 367706 / 25\% / 0\%                                                   \\
04/03/20      & 11351           & 11527               & 367706 / 25\% / 0\%                                                   & 367706 / 25\% / 0\%                                                   & 364148 / 24\% / 1\%                                                   \\
05/03/20      & 11306           & 11536               & 367706 / 25\% / 0\%                                                   & 367706 / 25\% / 0\%                                                   & 367706 / 25\% / 0\%                                                   \\
06/03/20      & 10943           & 11668               & 368892 / 25\% / 0\%                                                   & 370079 / 26\% / 1\%                                                   & 371265 / 26\% / 1\%                                                   \\
07/03/20      & 10876           & 11668               & 367706 / 25\% / 0\%                                                   & 367706 / 25\% / 0\%                                                   & 367706 / 25\% / 0\%                                                   \\
08/03/20      & 10809           & 11786               & 367706 / 25\% / 0\%                                                   & 367706 / 25\% / 0\%                                                   & 367706 / 25\% / 0\%                                                   \\
09/03/20      & 10742           & 11783               & 368892 / 25\% / 0\%                                                   & 367706 / 25\% / 0\%                                                   & 367706 / 25\% / 0\%                                                   \\
10/03/20      & 10538           & 11783               & 333308 / 13\% / 9\%                                                   & 370079 / 26\% / 1\%                                                   & 364148 / 24\% / 1\%                                                   \\
11/03/20      & 10334           & 11783               & 335680 / 14\% / 9\%                                                   & 367706 / 25\% / 0\%                                                   & 371265 / 26\% / 1\%                                                   \\
12/03/20      & 10040           & 11877               & 334494 / 14\% / 9\%                                                   & 365334 / 24\% / 1\%                                                   & 367706 / 25\% / 0\%                                                   \\
13/03/20      & 9108            & 11877               & 332122 / 13\% / 10\%                                                  & 298910 / 2\% / 19\%                                                   & 367706 / 25\% / 0\%                                                   \\
14/03/20      & 9268            & 11877               & 330936 / 12\% / 10\%                                                  & 306027 / 4\% / 17\%                                                   & 374823 / 27\% / 2\%                                                   \\
15/03/20      & 9428            & 11877               & 334494 / 14\% / 9\%                                                   & 296537 / 1\% / 19\%                                                   & 367706 / 25\% / 0\%                                                   \\
16/03/20      & 9588            & 11877               & 335680 / 14\% / 9\%                                                   & 372451 / 26\% / 1\%                                                   & 371265 / 26\% / 1\%                                                   \\
17/03/20      & 9285            & 11927               & 333308 / 13\% / 9\%                                                   & 303654 / 3\% / 17\%                                                   & 367706 / 25\% / 0\%                                                   \\
18/03/20      & 9088            & 11968               & 333308 / 13\% / 9\%                                                   & 298910 / 2\% / 19\%                                                   & 371265 / 26\% / 1\%                                                   \\
19/03/20      & 8063            & 12001               & 332122 / 13\% / 10\%                                                  & 296537 / 1\% / 19\%                                                   & 257394 / 13\% / 30\%                                                  \\
20/03/20      & 8284            & 12001               & 330936 / 12\% / 10\%                                                  & 298910 / 2\% / 19\%                                                   & 271628 / 8\% / 26\%                                                   \\
21/03/20      & 8172            & 12023               & 330936 / 12\% / 10\%                                                  & 298910 / 2\% / 19\%                                                   & 257394 / 13\% / 30\%                                                  \\
22/03/20      & 8059            & 12023               & 333308 / 13\% / 9\%                                                   & 306027 / 4\% / 17\%                                                   & 264511 / 10\% / 28\%                                                  \\
23/03/20      & 7946            & 12006               & 333308 / 13\% / 9\%                                                   & 294165 / 0\% / 20\%                                                   & 257394 / 13\% / 30\%                                                  \\
24/03/20      & 7848            & 12000               & 336866 / 14\% / 8\%                                                   & 301282 / 2\% / 18\%                                                   & 260953 / 11\% / 29\%                                                  \\
25/03/20      & 7735            & 11998               & 333308 / 13\% / 9\%                                                   & 296537 / 1\% / 19\%                                                   & 257394 / 13\% / 30\%                                                  \\
26/03/20      & 8451            & 11998               & 333308 / 13\% / 9\%                                                   & 301282 / 2\% / 18\%                                                   & 371265 / 26\% / 1\%                                                   \\
27/03/20      & 8949            & 12111               & 332122 / 13\% / 10\%                                                  & 298910 / 2\% / 19\%                                                   & 381940 / 30\% / 4\%                                                   \\
28/03/20      & 8761            & 12111               & 335680 / 14\% / 9\%                                                   & 296537 / 1\% / 19\%                                                   & 371265 / 26\% / 1\%                                                   \\
29/03/20      & 8574            & 12111               & 334494 / 14\% / 9\%                                                   & 301282 / 2\% / 18\%                                                   & 367706 / 25\% / 0\%                                                   \\
30/03/20      & 8386            & 12077               & 330936 / 12\% / 10\%                                                  & 294165 / 0\% / 20\%                                                   & 257394 / 13\% / 30\%                                                  \\
31/03/20      & 8529            & 12091               & 333308 / 13\% / 9\%                                                   & \textbf{303654 / 3\% / 17\%}                                          & 257394 / 13\% / 30\%                                                  \\ \hline
Total         & 294452          & 367706              &                                                                       &                                                                       &                                                                       \\ \hline

\end{tabular}

\end{center}

\end{table*}

\section{Experimental Results}
In this section, we present the empirical study of DTR along with the experimental setup and implementation details of DTR.

% \begin{table}
%  \caption{Sample table title}
%   \centering
%   \begin{tabular}{|p{2cm}|p{1cm}|p{1cm}|p{3.5cm}|p{3.5cm}|p{3.5cm}|}
    
%     Date     & Daily Actual     & {Daily Forecasted} & Reconciled Monthly Forecast \(\varepsilon\)=10\% & {Reconciled Monthly Forecast \(\varepsilon \)= 20\%} & {Reconciled Monthly Forecast \(\varepsilon\)=30\% }  \\
%     \hline
% 2020-02-01 &
%   71514 &
%   64553 &
%   2115056 \slash \hspace{0.3mm} -13\% \slash \hspace{0.3mm} 5\% &
%   2115056 \slash \hspace{0.3mm} -13\% \slash \hspace{0.3mm} 5\% &
%   2115056 \slash \hspace{0.3mm} -13\% \slash \hspace{0.3mm} 5\% \\
%     Dendrite & Input terminal  & $\sim$100     \\
%     Axon     & Output terminal & $\sim$10      \\
%     Soma     & Cell body       & up to $10^6$  \\
%     \bottomrule
%   \end{tabular}
%   \label{tab:table}
% \end{table} \(\varepsilon \)

\subsection{Dataset}
We perform experimental validation of our approach by using NIFTY 50 data set\cite{nifty50}. The NIFTY 50 index is the National Stock Exchange of India's benchmark broad based stock market index for the Indian equity market. NIFTY 50 stands for National Index Fifty, and represents the weighted average of the top 50 Indian company stocks in 17 sectors. This data set contains a time series starting from January 2000 till October 2021 containing indexes depicting the performance of the stock in form of Open price, Close price, High value per day, Low value per day, and Total volume traded per day. For most of the markets across the globe, the impact of COVID started after February 2020 which is why there is a steep decrease in the value of the time series in March 2020. The series starts recovering from April 2020. To capture this dynamic decrease, we have used the Open price index with the data starting from January 2019 till February 2020 as a training set and tested on March 2020. Given National Stock Exchange is only open on the weekdays from Monday to Friday, we interpolated the missing weekend dates. Fig. \ref{fig:rl_actual} shows high frequency data aggregated at the weekly level.   

\subsection{Performance Evaluation}
To evaluate the model performance, we have used the mean absolute percentage error (MAPE). This metric is calculated as:

\begin{equation}
MAPE = \frac{1}{n}{\sum_{t=1}^{n}\frac{|Y_t-\hat{Y_{t}}|}{|Y_t|} \ast 100 (\%)  },
\end{equation}
where \textit{n} is number of observations in time series, \( Y_t\) and \( \hat{Y_t}\) is observed and forecastes values at time step \textit{t}. 

We report the performance of different settings of proposed algorithms in Table \ref{table : results }.
For this experimentation, we have used fixed \(\epsilon\)-greedy policy along with 3 actions defined based on Section 3.2.  The results in all experiments are given by “ Reconciled Monthly forecast/ MAPE of reconciled forecast wrt monthly actual(\( MAPE_{rec}\) )/ Percentage improvement over monthly forecast without reconciliation (\( \%_f)\) ”. As can be seen, the proposed DTR clearly outperform the both high frequency aggregated forecast as well as low frequency without reconciliation forecast. The framework starts with aggregated high frequency forecast and based on the actual seen per day changes its estimate for low frequency forecast. The results reveal how the DTR framework adapts to high frequency actual based on different parameter initialisation. If there is unexpected increase/decrease in  daily values between the  days, the framework increases/decreases its estimated low frequency forecast which is shown in Table \ref{table : results }. On Day 13, there is a huge decrease in high frequency actual going from  (\ 10040\hspace{0.5mm} -> 9108\hspace{0.5mm} )\ which makes the framework to adjust estimated low frequency forecast going from  (\ 365334\hspace{0.5mm} -> 298910\hspace{0.5mm} )\ based on  5\% tolerance and \(\varepsilon\) as 20 \% . The reconciled forecast are highly dependent on tolerance level. Tolerance level parameter is problem specific and should be adjusted based on the business needs. We have tested our framework on values of \(\varepsilon\)  at (\ 10\%, 20\%, 30\% )\ of average high frequency forecast value which is (\ 36736, 73473, 110209 )\ respectively.

\subsection{Hyperparameter Analysis}
The proposed framework has two hyper-parameters \( \epsilon \) and \( \varepsilon \). \( \epsilon \) is used to balance between exploration and exploitation which helps in selecting the action with the highest estimated reward most of the time. \( \varepsilon \) is a parameter to
adjust the strictness of the agent. Table 3 contains the effects of \( \epsilon\) and \(\varepsilon \) on the agent’s performance. This provides a user the flexibility to change the agent’s behavior according to their requirement. Smaller \( \epsilon\) will rely on greedy action that agent believes has the best long-term effect. A higher \( \epsilon\) will help agent to explore more. Similarly smaller \( \varepsilon\) will lead to adjustment in forecast near the mean value of forecast only whereas a higher \( \varepsilon\) will help agent to model high variance changes also. 

\begin{table}[ht]
\caption{Effect on agent by varying  \(\varepsilon\) and  \(\epsilon\) for NIFTY 50 dataset }
\label{table : hyper }

\centering

\begin{tabular}{|c|c|c|c|}
\hline
\textbf{tolerance} & \textbf{epsilon} & \textbf{\( MAPE_{rec}\)} & \textbf{\( \%_f\)} \\ \hline
10\%               & 0.05             & 13\%                     & 9\%                \\
20\%               & 0.05             & 3\%                      & 17\%               \\
30\%               & 0.05             & 13\%                     & 30\%               \\
10\%               & 0.1              & 15\%                     & 8\%                \\
20\%               & 0.1              & 5\%                      & 16\%               \\
30\%               & 0.1              & 5\%                      & 24\%               \\
10\%               & 0.2              & 16\%                     & 7\%                \\
20\%               & 0.2              & 6\%                      & 15\%               \\
30\%               & 0.2              & 4\%                      & 23\%               \\ \hline
\end{tabular}

\end{table}

\section{Conclusion}
In this paper we present a Dynamic Temporal Reconciliation
(DTR) framework by incorporating a dynamic adjustment layer into
traditional Reconciliation algorithms. The integration allows us to better model the complex changes in high frequency forecast and adjust long term estimates accordingly. Therefore, DTR provides a powerful tool to handle incomplete low frequency data for reconciliation tasks based on high frequency actual and is well equipped to overcome the deficiencies of the existing reconciliation methods. For reconciliation, we propose a Reinforcement learning framework which is formulated as a MDP. This has been solved using a Time Differenced Reinforcement learning (TDRL) approach with custom action design and reward function. Proposed framework is data-driven and can be customised based on business needs. We examine the framework on NIFTY 50 dataset\cite{nifty50}, and DTR framework has demonstrated superior performance over baseline forecasts. Although we introduce DTR in an environment with only 3 actions and fixed \( \epsilon\)-greedy policy, the framework can be customised by changing numbers of actions designed and by learning a dynamic policy function. 
%Bibliography
\bibliographystyle{unsrt}  
\bibliography{references}

\end{document}